\documentclass[10pt,journal]{IEEEtran}

\usepackage{url}
\usepackage{epsfig}
\usepackage{subfigure}
\usepackage{graphicx}
\usepackage{algorithm}
\usepackage{algorithmic}
\usepackage{amssymb}
\usepackage{mathrsfs} 
\usepackage{amsmath,bm}
\usepackage{url}
\usepackage{color}
\usepackage{caption}
\usepackage{booktabs}
\usepackage{multirow}
\usepackage{array}
\usepackage[colorlinks=true,linkcolor=blue,anchorcolor=blue,citecolor=blue]{hyperref}

\newcommand{\argmax}{\mathop{\arg\!\max}}

\def\comment#1{{}}
\def\eg{{\em e.g.}}
\def\ie{{\em i.e.}}
\def\etal{{\em et al.}}

\begin{document}
%
\title{Geometric Hypergraph Learning for Visual Tracking}

\author{Dawei Du, Honggang Qi, Longyin Wen, Qi Tian, Qingming Huang, and Siwei Lyu
\thanks{Dawei Du, Honggang Qi, and Qingming Huang are with the School of Computer and Control Engineering, University of Chinese Academy of Sciences, Beijing, China. E-mail:cvdaviddo@gmail.com}
\thanks{Qi Tian is with the Department of Computer Science, the University of Texas at San Antonio (UTSA), USA.}
\thanks{Longyin Wen and Siwei Lyu are with the Computer Science Department, University at Albany, SUNY, Albany NY, USA. Siwei Lyu is also with School of Computer and Information, Tianjin Normal University, Tianjin, China.}
}

%

\maketitle

\begin{abstract}
  Graph based representation is widely used in visual tracking field by finding correct correspondences between target parts in consecutive frames. However, most graph based trackers consider pairwise geometric relations between local parts. They do not make full use of the target's intrinsic structure, thereby making the representation easily disturbed by errors in pairwise affinities when large deformation and occlusion occur. In this paper, we propose a geometric hypergraph learning based tracking method, which fully exploits high-order geometric relations among multiple correspondences of parts in consecutive frames. Then visual tracking is formulated as the mode-seeking problem on the hypergraph in which vertices represent correspondence hypotheses and hyperedges describe high-order geometric relations. Besides, a confidence-aware sampling method is developed to select representative vertices and hyperedges to construct the geometric hypergraph for more robustness and scalability. The experiments are carried out on two challenging datasets (VOT2014 and Deform-SOT) to demonstrate that the proposed method performs favorable against other existing trackers.
\end{abstract}


\ifCLASSOPTIONpeerreview
\begin{center} \bfseries EDICS Category: 3-BBND \end{center}
\fi
%
\IEEEpeerreviewmaketitle

\section{Introduction}
Visual tracking has attracted much research interest in computer vision field, because it is a critical step of various applications, including video surveillance, sport analysis, auto-drive car, etc. Despite having achieved promising progress over the past decade, it still remains very challenging for designing a robust tracker that can handle appearance changes caused by various critical situations, such as large deformation, illumination variation, partial and full occlusion, and background clutter. In particular, the deformation and occlusion are the two most notable challenges that degrade tracking performances.

For tracking scenarios where the target appearance is relatively stable, methods based on global appearance models can achieve satisfactory performances~\cite{kalal2010pn, hare2011struck, DBLP:journals/tsmc/WangCX11, zhang2014fast, DBLP:conf/bmvc/DanelljanHKF14, DBLP:journals/pami/HenriquesC0B15}. However, if large deformation and occlusion happen, such approaches usually fail to track the target robustly. To counter this problem, many approaches based on models of local parts have received more attention~\cite{wang2011superpixel, oron2012locally, DBLP:conf/cvpr/JiaLY12, DBLP:journals/tcyb/WangY14b, DBLP:journals/tcyb/YuYWH15}. Moreover, several different methods to represent the target geometric structure have been proposed, such as structural Support Vector Machine (SVM)~\cite{yao2013part}, Markov Random Field (MRF)~\cite{ren2007tracking, DBLP:conf/eccv/HongWMPT14}, keypoint constellation~\cite{DBLP:conf/wacv/NebehayP14, nebehay2015clustering}, and graph model~\cite{DBLP:conf/accv/WangN12, cai2014robust}. However, most approaches consider pairwise relations between target parts are easily disturbed by errors in pairwise affinities, rendering difficulties to well preserve the geometric structure underlying the target representation.
\begin{figure}[t]
\centering
\includegraphics[width=.75\linewidth]{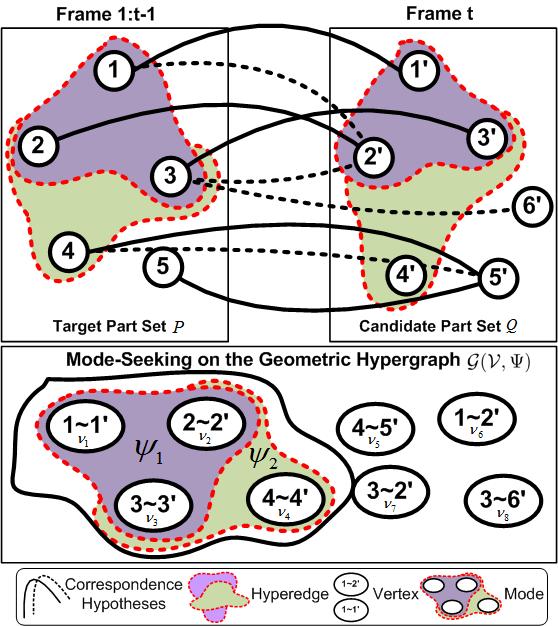}
\caption{In the top, between the target part set $P$ and the candidate part set $Q$, the correspondence hypotheses are generated and constrained by the high-order relations among them. In the bottom, the geometric hypergraph $\mathcal{G}$ is constructed based on $P$ and $Q$. Then the mode is extracted by searching $\mathcal{G}$. For clarity, just a few vertices and hyperedges are shown.}
\label{fig_geometric_model}
\end{figure}

In this paper, we present a novel Geometric hyperGraph Tracker (GGT) to handle the visual tracking task, especially for the deformable ones. Different from the previous works considering pairwise geometric relations between local parts, our method exploits high-order relations among more than two correspondences based on the {\em geometric hypergraph}. Specifically, the geometric hypergraph is constructed and learned based on the target part set and the candidate part set\footnote{The candidate part set $Q$ consists of candidate parts extracted from the searching area in the current frame $t$. We employ the target part set $P$ as the part representation of the target, which is consisted by the target parts up to the previous frame $t-1$.}. Then we generate possible correspondences between parts in the two sets, which are defined as {\em correspondence hypotheses}. In Fig.~\ref{fig_geometric_model}, we give a schematic diagram of constructing the hypergraph $\mathcal{G}(\mathcal{V}, \Psi)$, where vertices $\mathcal{V}$ encode correspondence hypotheses, and hyperedges $\Psi$ encode high-order geometric relations among several correspondence hypotheses. Thus the geometric structure of the target can be effectively characterized by the hypergraph, rendering more discriminative power to extract the common appearance and geometric property of correspondences from noises. Moreover, correspondence hypotheses can form a set of modes where a large number of hyperedges are involved with high confidence; while other false correspondences have very few hyperedges with low confidence. For easier reading, we first define the structural correspondence mode as:

{\noindent \textbf{Definition 1.}} \textit{The structural correspondence mode is a group of reliable correspondences between target parts with similar appearance and consistent geometric structure that are inter-connected with the local maximum of overall confidences on the geometric hypergraph.}

The present work makes the following contributions:
\begin{itemize}
  \item The geometric hypergraph is used to represent the target, which fully exploits high-order geometric relations among correspondence hypotheses in consecutive frames.
  \item The confidence-aware sampling method is proposed to approximate the geometric hypergraph, which not only alleviates sensitivity to noises, but also is scalable to the large scale hypergraph. Thus we seek structural correspondence modes on the hypergraph by the pairwise coordinate update method in~\cite{liu2012dense} efficiently.
  \item Our method is compared to existing methods on the VOT2014 dataset and Deform-SOT dataset. The experimental results demonstrate the effectiveness and robustness of the proposed model.
\end{itemize}

The rest of the paper is organized as follows. In Section \ref{sec_related_work}, we review relevant previous works. The methodology is described in Section \ref{sec_our_method} and the model optimization is presented in Section \ref{sec_optimization}. In Section \ref{sec_experiment}, we evaluate the proposed algorithm on two tracking datasets compared to other existing methods. Then we conclude the paper with discussions on future works in Section \ref{sec_conclusion}.
\section{Related Works}\label{sec_related_work}
Tracking methods based on modeling relations between target parts have been shown to be less susceptible to the problem posed by object deformation and occlusion. Recently, many works have focused on how to incorporate geometric information as an important clue to facilitate visual tracking.

{\noindent \textbf{Keypoint Based Tracking Methods.}} The keypoint based trackers use the displacements of target parts to vote for the target center in consecutive frames to consider the geometric structure. Hare \etal~\cite{DBLP:conf/cvpr/HareST12} combine the feature matching, learning, and object pose estimation into a coherent structured output learning framework, resulting in learning for real-time keypoint-based object detection and tracking. Yang \etal~\cite{DBLP:journals/ivc/YangLY13} propose a visual tracking algorithm by incorporating the SIFT features from the interest points to represent appearance and exploiting their geometric structures, where a structured visual dictionary is learned to enhance its discriminative strength between the foreground object and the background. Yi \etal~\cite{DBLP:conf/iccv/YiJHCC13} propose a tracking method using ``motion saliency'' and ``descriptor saliency'' of local features and performs tracking based on generalized Hough transform. The tracking result is obtained by combining the results of each local feature of the target and the surroundings with generalized Hough transform voting. Guo \etal~\cite{DBLP:journals/cviu/GuoCTLLL14} formulate the task of tracking and recognition as a maximum posteriori estimation problem under the manifold representation learnt from collections of local features with preserving local appearance similarity and spatial structure. Nebehay and Pflugfelder~\cite{DBLP:conf/wacv/NebehayP14} develop a keypoint-based tracking method in a combined matching-and-tracking framework, where each keypoint casts votes for the object center. Moreover, an improved algorithm in~\cite{nebehay2015clustering} employs geometric dissimilarity measure to separate inlier correspondences from outliers by considering both static and adaptive correspondences. In~\cite{DBLP:journals/cviu/BouachirB15}, keypoints are considered as elementary predictors localizing the target in a collaborative search strategy, where the persistence, the spatial consistency, and the predictive power of a local feature are used to to measure the most reliable features for tracking. However, keypoint based trackers focus on modeling the displacements between parts and the corresponding target center, which are insufficient to exploit relations between local parts fully for geometric structure representation.

{\noindent \textbf{Part Based Tracking Methods.}} To better solve the shape deformation and partial occlusion issue, part based methods are gaining popularity in visual tracking. Wen \etal~\cite{DBLP:conf/accv/WenCDLL14} present a discriminative learning method to infer the position, shape and size of each part, using the Metropolis-Hastings algorithm integrated with an online SVM. Wang and Nevatia~\cite{DBLP:conf/accv/WangN12} propose to track non-rigid objects with multiple related parts and model tracking as Dynamic Bayesian Network, where the spatial relations among parts are formulated probabilistically. Improved from~\cite{hare2011struck}, Yao \etal~\cite{yao2013part} introduce a part-based tracking algorithm with online latent structured learning, and use a global object box and a small number of part boxes to approximate the irregular object, to reduce the amount of visual drift. Cehovin \etal~\cite{cehovin2013robust} employ a global representation to probabilistically models target's global visual properties. Meanwhile, the low-level patches are constrained and updated with the global model during tracking. A dynamic structure graph based tracker is used in~\cite{cai2014robust} to formulate the tracking problem as subgraph matching between the geometric structure graph of the target and that of the candidate target proposals graph. Nam \etal~\cite{DBLP:conf/eccv/NamHH14} use a new graphical model to adapt sequence structure and propagate the posterior over time, where each vertex has a single outgoing edge but may has multiple incoming edges. Hong \etal~\cite{DBLP:conf/eccv/HongWMPT14} propose a MRF-based tracker to consider geometric structure by the hierarchical appearance representation, which exploits shared information across multi-level quantization of an image space, \ie, pixels, superpixels and bounding boxes. In~\cite{DBLP:conf/cvpr/LiuWY15}, a real-time tracking method is proposed based on parts with multiple correlation filters, in which the Bayesian inference framework and a structural constraint mask are adopted to handle various appearance changes. However, the existing part based methods give less consideration to exploit high-order geometric relations among target parts for more robustness.

{\noindent \textbf{Segmentation Based Tracking Methods.}} The segmentation based methods consider the geometric information by finding out the precise location of each pixel in the target. Based on the generalized Hough-transform, Godec \etal~\cite{DBLP:conf/iccv/GodecRB11} develop an improved online Hough Forests and couple the voting based detection and back-projection with a rough segmentation based on GrabCut. Duffner and Garcia present a pixel-based non-rigid object tracking method in~\cite{DBLP:conf/iccv/DuffnerG13}, which consists of a generalized Hough transform with pixel-based descriptors based detector and a probabilistic segmentation method based on a global model for foreground and background. Recently, Wen \etal~\cite{wen2015jots} develop a joint online tracking and segmentation algorithm, which integrates the multi-part tracking and segmentation into a unified energy optimization framework.
\section{Methodology}\label{sec_our_method}
\begin{figure*}[tp]
\centering
\includegraphics[width=.95\linewidth]{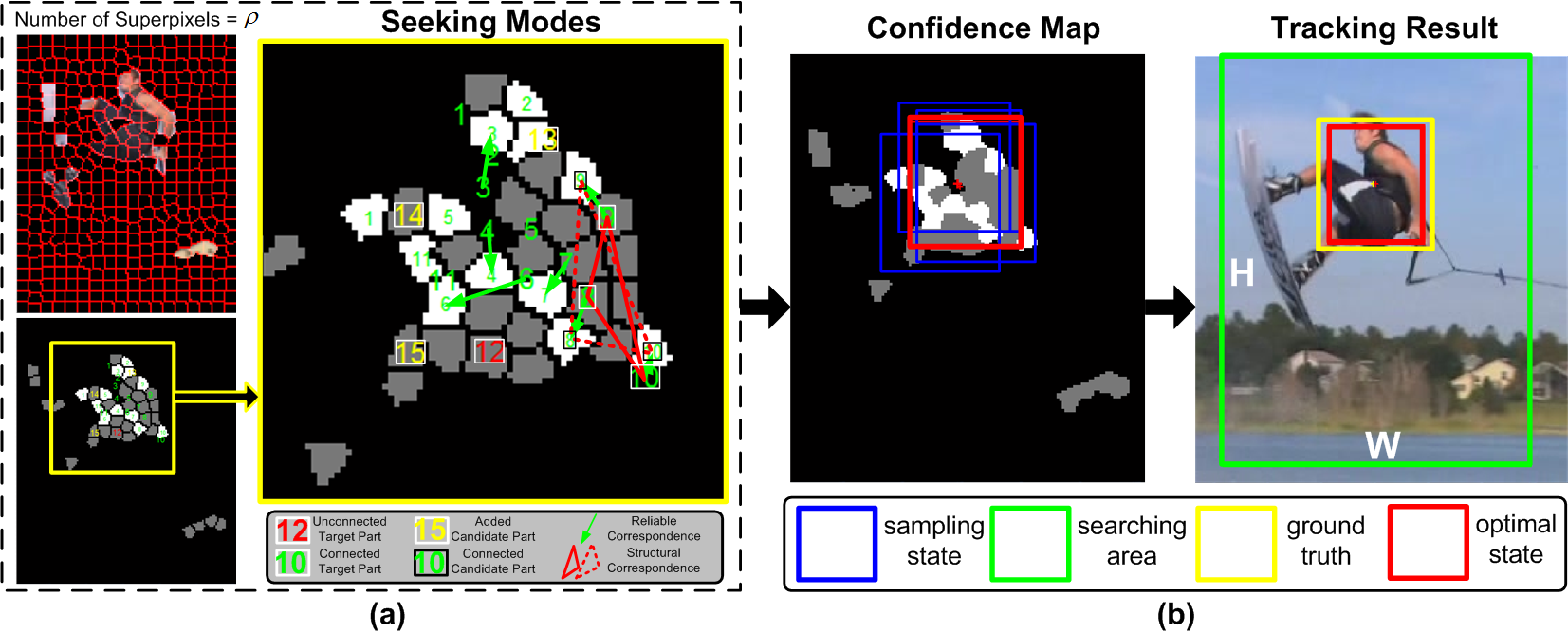}
\caption{Tracking process on the \textit{waterski} sequence. (a) Given the candidate part set $Q$, we aim to find their reliable correspondence with the target part set $P$. This is done by seeking structural correspondence modes with tolerance of deformation and scale change. For example, the green arrows in the figure indicate the displacement between parts in $P$ and $Q$. Then reliable target parts can be determined (\eg, part $8$, $9$, $10$), and the geometric hypergraph is updated incrementally. (b) The confidence map is constructed based on the reliable target part set, and the tracking result is output by uniform sampling on the confidence map.}
\label{fig_example}
\end{figure*}

We first introduce terms and notations to be used in the sequel. We denote the order of hypergraph as $k$. The hypergraph is a generalization of a graph in which an edge (hyperedge, strictly speaking) can connect more than $k(k\geq3)$ vertices, while a graph has its edges connecting $2$ vertices. The unconnected graph is a graph without edges between vertices.

Although our method is related with three previous works including SPT~\cite{wang2011superpixel}, DGT~\cite{cai2014robust} and TCP~\cite{li2015online}, there are significant differences between our method and them, which are concluded below.
\begin{itemize}
  \item Though both our method and SPT use superpixel representation, our method uses a more effective representation of the geometric information of target parts, which leads to improved performance on more complex scenes. When the hypergraph degenerates into an unconnected graph ($k=1$), SPT can be regarded as a special case of the proposed algorithm.
  \item DGT uses a graph to exploit pairwise geometric relations between neighboring parts. On the contrary, our method employs a hypergraph that considers high-order geometric relations among correspondence hypotheses to exploit abrupt deformation, motion change and target context better. When the hypergraph reduces to a normal graph ($k=2$), DGT can be regarded as a special case of the proposed algorithm.
  \item TCP mainly exploits temporal high-order relations among different parts in consecutive frames, ignoring the geometric structure information of local parts spatially. In contrast, our method focuses on modeling the spatial high-order relations among correspondence hypotheses. Besides, the temporal relations of parts are also considered when updating the hypergraph.
\end{itemize}

In this work, the tracking problem is formulated as the mode-seeking problem on the geometric hypergraph. In Section~\ref{sec_graph_construction}, we construct the hypergraph based on the target part set and candidate part set. Then we give the detailed formulation in Section~\ref{sec_hyper_graph} and define corresponding confidence measure in Section~\ref{sec_confidence_calculation}.
\subsection{Geometric Hypergraph}\label{sec_graph_construction}
The superpixel representation is more flexible for the deformable target compared to the holistic representation, however has low discriminative power because of small size. Therefore we construct a geometric hypergraph to alleviate the problem with geometric constraints. Given the annotated bounding box in the first frame, the target part set $P$ is first initialized, and the candidate part set $Q$ is determined by the coarse labeling of superpixels in the rest frames\footnote{Similar as~\cite{cai2014robust}, we first use the SLIC algorithm~\cite{achanta2012slic} to over-segment the searching area of the target into multiple parts (superpixels), and employ the Graph Cut method~\cite{DBLP:journals/pami/BoykovK04} to coarsely separate the foreground parts from the background, as shown in the top-left of Fig.~\ref{fig_example}(a).}. Based on $P$ and $Q$, we construct the vertex set ${\cal V}$ and hyperedge set $\Psi$ of the geometric hypergraph ${\cal G}$ as
\begin{equation}
\label{equ_geometric_hypergraph}
\left\{
\begin{aligned}
&{\cal V} = \{\nu_i\}_{i=1}^{N}=\{ p\sim q |\forall p\in P, q\in Q: d_{E}(p,q) \leq \tau_d \}\\
&\Psi = \{\psi|\forall \nu_i,\nu_j\in\psi: \nu_i\cap \nu_j=\emptyset\}\\
\end{aligned}
\right.
\end{equation}
where $N$ is the number of vertices. $\nu_i$ and $\nu_j$ are the $i$-th and $j$-th vertex in hyperedge $\psi = \{\nu_1, \cdots, \nu_k\}$ without conflicts or duplicates. $k$ is the order of hypergraph. $d_{E}(p,q)$ is the Euclidean distance between the centers of parts $p$ and $q$ in image plane. The distance threshold is set as $\tau_d=3\sqrt{W\cdot H/\rho}$, where $\rho$ is the number of superpixels in the searching area with width $W$ and height $H$ in the current frame, as shown in Fig.~\ref{fig_example}.
\subsection{Formulation}\label{sec_hyper_graph}
As analyzed in the introduction, multiple correspondences with similar structural geometric properties form a set of structural correspondence modes. By measuring the overall confidence of modes, the tracking problem is formulated as
\begin{align}
\label{equ_hypergraph_model}
&\mathcal{D}^\ast=\argmax_{\mathcal{D}}\Omega\big(\mathcal{D}\big),\nonumber\\
&\text{s.t.}\quad \mathcal{D}\subset {\cal G},|\mathcal{D}|=\kappa,
\end{align}
where $\mathcal{D}$ is the structural correspondence mode including $\kappa$ number of vertices. $\Omega(\mathcal{D})$ is the confidence measure function reflecting the confidence distribution in the mode $\mathcal{D}$, which is described as follows.
\subsection{Confidence Measure}\label{sec_confidence_calculation}
We design two terms to encode both the association confidence $\Gamma$ of vertices and the geometric confidence $\Xi$ among them, \ie,
\begin{equation}
\label{equ_association_confidence}
\begin{aligned}
\Omega\big(\mathcal{D}\big) = \omega_1\cdot\underbrace{\sum_{\nu\in\mathcal{N}(\mathcal{D})}{\Gamma(\nu)}}_{\text{Association Confidence}} + \omega_2\cdot\underbrace{\sum_{\psi\in\mathcal{E}(\mathcal{D})}{\Xi(\psi)}}_{\text{Geometric Confidence}},
\end{aligned}
\end{equation}
where $\mathcal{N}(\mathcal{D})$ and $\mathcal{E}(\mathcal{D})$ denote the vertex set and the hyperedge set of mode $\mathcal{D}$, respectively. $\omega_1$ and $\omega_2$ are the balancing factors of the two terms.

{\noindent \textbf{Association Confidence.}} The association confidence $\Gamma(\nu)$ encodes the probability of two parts in vertex $\nu\in\mathcal{N}(\mathcal{D})$ belonging to the same class, which is defined as
\begin{equation}
\label{equ_self_circle}
\begin{aligned}
\Gamma(\nu) = \exp\big[-\frac{1}{\sigma_\nu^2}d_{\chi}(p,q)\big],
\end{aligned}
\end{equation}
where $d_{\chi}(p,q)$ is the $\chi2$ distance between the appearance feature of the two parts, \ie, $\nu=\{p,q\}$, $p\in P$ and $q\in Q$. In the experiment, the appearance feature is concatenated by HSV color feature and LBP texture. $\sigma_\nu$ is the scaling parameter to measure the importance of appearance similarity.

{\noindent \textbf{Geometric Confidence.}} The geometric confidence $\Xi(\psi)$ describes the geometric relation among correspondence hypotheses in hyperedge $\psi$. If the order of graph $k\geq3$, it is a hypergraph describing high-order geometric relations among correspondence hypotheses. When reducing the order of graph, it degenerates into a graph to consider pairwise geometric relations, and an unconnected graph ignoring geometric relations. Therefore, for different order of hypergraph case, we have different calculation of geometric confidence, which is discussed as follows.
\subsubsection{Unconnected Graph}
If $k=1$, the hypergraph becomes an unconnected graph, \ie, $\Xi(\psi) = \emptyset$. Thus visual tracking only depends on the association confidence $\Gamma(\nu)$ without any geometric structural constraints. Similar to SPT~\cite{wang2011superpixel}, it is actually a part-based template matching method. The appearance information encoded in $\Gamma(\nu)$ is usually weak especially for small superpixels, resulting in worse performance in the scenarios with complex appearance variation.
\subsubsection{Graph}
If $k=2$, the geometric information encoded in $\Xi(\psi)$ provides complementary pairwise geometric information of edge $\psi=\{\nu_1,\nu_2\}$ besides appearance than SPT~\cite{wang2011superpixel} does. Thus DGT~\cite{cai2014robust} falls into this category. The pairwise similarity to compare two correspondence hypotheses is calculated as
\begin{equation}
\label{equ_2_affinity}
\Xi(\psi)=\exp\big[-\frac{1}{\sigma_\psi^2}||\vec{L}(p_1,p_2)-\vec{L}(q_1,q_2)||_2\big],
\end{equation}
where $p_1$ and $p_2$ denote the parts in target part set $P$, $q_1$ and $q_2$ the parts in candidate part set $Q$. $\vec{L}(\cdot,\cdot)$ measures the consistency of the two \textit{supporters}, which is calculated as the location displacement of two neighboring correspondence hypotheses, as shown in Fig.~\ref{fig_triangle_model}(a). $\sigma_\psi$ is the scaling parameter to measure the importance of geometric constraint.
\subsubsection{Hypergraph}
\begin{figure}[t]
\centering
\includegraphics[width=.85\linewidth]{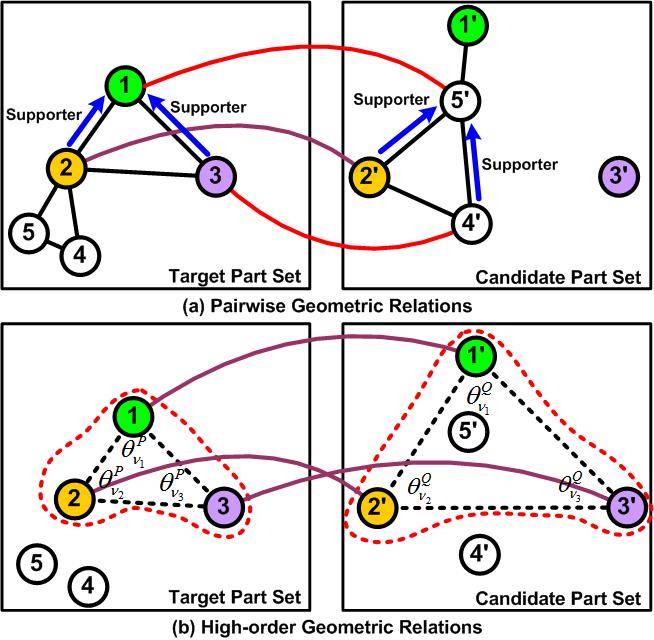}
\caption{(a) Pairwise geometric relations. When large scale changes occur, the wrong correspondences between target parts (\eg, $1\sim5'$ and $3\sim4'$) are easily conducted since the supporters (shown in blue arrow) are no longer reliable. (b) High-order geometric relations. Different from the pairwise measure, the angles of the triplet hypotheses (\eg, $\theta_{\nu_1}^P\sim\theta_{\nu_1}^Q$, $\theta_{\nu_2}^P\sim\theta_{\nu_2}^Q$, and $\theta_{\nu_3}^P\sim\theta_{\nu_3}^Q$) are invariant to large scale changes, leading to correct correspondences.}
\label{fig_triangle_model}
\end{figure}

As shown in Fig.~\ref{fig_triangle_model}(a), the supporters provide pairwise relation measure that are restricted to distances, making it hard to handle large scale change and conducting wrong correspondences between target parts. On the contrary, as shown in Fig.~\ref{fig_triangle_model}(b), we exploit the angle information of triplets of correspondence hypotheses to achieve scale invariance, leading to more correct associations between target parts. For example, three correspondence hypotheses form two triangles ($\bigtriangleup_{1,2,3}$ and $\bigtriangleup_{1',2',3'}$). Although the target scale changes drastically, the angles (high-order geometric relations) remain more stable compared to the relative displacement between parts (pairwise geometric relations). For better understanding, a real example of structural correspondence in \textit{waterski} is shown in Fig.~\ref{fig_geometric_model}(b). Similar to~\cite{DBLP:conf/cvpr/LeeCL11}, the geometric confidence is calculated by comparing corresponding angles, as
\begin{equation}
\label{equ_3_affinity}
\Xi(\psi)=\exp\big[-\frac{1}{\sigma_\psi^2}\sum_{i=1}^{3}{|\sin(\theta_{\nu_i}^P)-\sin(\theta_{\nu_i}^Q)|}\big],
\end{equation}
where $\theta_{\nu_i}^P$ and $\theta_{\nu_i}^Q$ denote angles of parts related to vertex $\nu_i$ in the set $P$ and $Q$, respectively.
\section{Optimization}\label{sec_optimization}
Given the geometric hypergraph $\mathcal{G}$, the mode-seeking problem is solved by searching $\mathcal{G}$ (Section~\ref{sec_search}). Before that, We propose the confidence-aware sampling technique to improve the effectiveness of the proposed method (Section~\ref{sec_sampling}).
\subsection{Mode-Seeking Problem} \label{sec_search}
Since the maxima of~\eqref{equ_hypergraph_model} corresponds to a structural correspondence mode, we fully search the hypergraph $\mathcal{G}$ by setting each vertex $\nu^\star$ in the hypergraph as a starting point. Let $\mathcal{D}_{\nu^\star}$ be the mode with vertex set $\mathcal{N}(\mathcal{D}_{\nu^\star})$ and hyperedge set $\mathcal{E}(\mathcal{D}_{\nu^\star})$. Let $\mathcal{P}\in\textbf{R}^N$ be the vector containing the probability of each vertex in the hypergraph belonging to mode $\mathcal{D}_{\nu^\star}$, \ie, if $v\in\mathcal{N}(\mathcal{D}_{\nu^\star})$, $\mathcal{P}_{v}>0$; otherwise, $\mathcal{P}_{v}=0$. $N$ is the number of vertices in the hypergraph. Combined with~\eqref{equ_association_confidence}, the problem in~\eqref{equ_hypergraph_model} is cast as optimizing the probability vector $\mathcal{P}$ and further rewritten as
\begin{align}
\label{equ_mode}
\mathcal{P}^\ast = &\argmax_{\mathcal{P}_{v}:v\in{\cal N}(\mathcal{D}_{\nu^\star})}\big(\sum_{v\in{\cal N}(\mathcal{D}_{\nu^\star})}{\Gamma(v)\mathcal{P}_{v}} + \sum_{e\in\mathcal{E}(\mathcal{D}_{\nu^\star})}{\Xi(e)\prod_{v\in e} \mathcal{P}_{v}}\big),\nonumber\\
&\text{s.t.}\quad \sum_{v\in{\cal V}}\mathcal{P}_{v}=1,\mathcal{P}_{v}\in\{0,\mu\},\frac{1}{\mu}\geq k+1,
\end{align}
in which the first term in the objective function penalizes the inclusion of vertices corresponding to less association confidence indicated by a lower $\Gamma(v)$, and the second term encourages the inclusion of hyperedges in the mode with larger geometric confidence $\Xi(e)$. Essentially, this is a NP-hard combinatorial optimization problem. To solve this problem, the constraint $\mathcal{P}_{v}\in\{0,\mu\}$ is relaxed to $\mathcal{P}_{v}\in[0, \mu]$, where $\mu\leq1$ is a constant. Let the number of vertices in $\mathcal{D}_{\nu^\star}$ be $m$, the mode contains at least $m=[\frac{1}{\mu}]$ number of vertices when keeping the constraint $\sum_{v\in{\cal V}}\mathcal{P}_{v}=1$. To avoid the degeneracy problem, we require the minimal vertices in a mode satisfying the constraint $\frac{1}{\mu}\geq k+1$ to guarantee adequate structural correspondences included in one mode. Then the pairwise coordinate updating method~\cite{liu2012dense} is used to solve the problem in~\eqref{equ_mode} effectively. Refer to~\cite{liu2012dense} for more details about the optimization strategy.
\subsection{Confidence-aware Sampling} \label{sec_sampling}
Suppose that the target part set and candidate part set consist of $n$ parts, there are at most $n^2$ correspondence hypotheses. For the $k=3$ case, the size of the resulting full-affinity hyperedges will be $\binom{n^2}{3}$, of order $O(n^6)$, which demands a huge amount of memory. It becomes than necessary to further reduce the computational complexity by introducing a sparse hypergraph structure with significant hypotheses. To this end, we develop a confidence-aware sampling method as follows.
\begin{enumerate}
  \item Firstly, we reduce the number of vertices deterministically by introducing two thresholds. We assume target parts move smoothly in consecutive frames, which means that the appearances change little in a very short time interval. To remove noises, for each target part $p\in P$, we only consider a few correspondence hypotheses with at most $\varsigma=5$ number of highest association confidence larger than an appearance threshold $\epsilon_a$.
  \item Secondly, the number of hyperedges is greatly decreased probabilistically. Based on a simple assumption that a vertex with higher association confidence has a higher possibility of being reliable correspondence, we sample more hyperedges around the vertex with higher association confidence. Specifically, starting from each vertex $\nu$ in the reduced vertex set, we sample $\eta=[\hat{\Gamma}(\nu)\cdot N_\nu]$ number of hyperedges comprising three vertices without conflicts. We regard the normalized confidence $\hat{\Gamma}(\nu)$ as the sampling probability, and the constant $N_\nu$ as the maximal number of sampled hyperedges for each vertex.
\end{enumerate}
Different from other MRF or graph based approaches considering pairwise relations between the nearest neighboring vertices, we sample hyperedges randomly without distance constraints to exploit the geometric information fully, so that the hypergraph is spanned globally over all correspondence hypotheses. The additional benefit is that we can consider context information between target parts and background parts for more robustness.

Based on the confidence-aware sampling method, we sample vertices and hyperedges of $\mathcal{G}$, obtaining a approximate geometric hypergraph $\mathcal{G}^\ast$. Then we directly perform mode-seeking on $\mathcal{G}^\ast$ instead of $\mathcal{G}$. Specifically, the reduced vertex set $\mathcal{V}^\ast$ and hyperedge set $\Psi^\ast$ of $\mathcal{G}^\ast$ are given as
\begin{equation}
\label{equ_sampling}
\left\{
\begin{aligned}
&\mathcal{V}^\ast=\{\nu|\forall \nu\in\mathcal{V}:\Gamma(\nu)\geq\epsilon_a, |\mathcal{V}^p|\leq\varsigma\},\\
&\Psi^\ast = \{\psi|\forall \nu\in\mathcal{V}^\ast, \nu_i,\nu_j\in\psi: |\Psi^\nu|\leq \eta, \nu_i\cap\nu_j=\emptyset\}\\
\end{aligned}
\right.
\end{equation}
where $|\mathcal{V}^p|$ denotes the number of vertices including part $p$, and $|\Psi^\nu|$ denotes the number of hyperedges including vertex $\nu$. The sampling scheme ensures finding enough relevant correspondence hypotheses. Moreover, it decreases the number of vertices from $n^2$ to at most $n\varsigma$ and the number of hyperedges from $\binom{n^2}{3}$ to at most $n\varsigma\eta$, which removes more than redundant $90\%$ of vertices and hyperedges in $\mathcal{G}$ empirically.
\section{Tracking}
\subsection{Extracting Reliable Target Parts}
Given the optimized probability vector $\mathcal{P}^\ast$, we can determine the vertices belonging to the corresponding mode $\mathcal{D}$, \ie, $\mathcal{D}=\{\nu|\forall \nu\in\mathcal{V}:\mathcal{P}_{\nu}>0\}$. Since the hypergraph is searched starting from each vertex, one vertex may appear in multiple modes. The conflicts involved in the modes should be removed to find reliable target parts $\mathcal{S}$, and the whole procedure is summarized in Algorithm~\ref{alg_post_processing}.
\begin{algorithm}[t]
\caption{Extracting Reliable Target Parts}
\label{alg_post_processing}
\small {
\begin{algorithmic}[1]
\REQUIRE structural correspondence mode set $\textbf{D}$
\ENSURE reliable target part set $\mathcal{S}$
\STATE Sort the mod set $\textbf{D}=\{\mathcal{D}_{1},\cdots,\mathcal{D}_{N}\}$ based on the confidence values $\{\Omega(\mathcal{D}_1),\cdots,\Omega(\mathcal{D}_N)\}$ in descending order
\STATE Initialize the mode set without conflicts $\textbf{D}^\ast = \emptyset$
\FOR{each non-empty mode $\mathcal{D}_{i}\in\textbf{D},\mathcal{D}_{i}\neq\emptyset$}
\IF{no intersections with all members in the set, \ie, $\forall j, \mathcal{D}_{\nu_j}^\ast\in\textbf{D}^\ast: \mathcal{D}_{i}\cap\mathcal{D}^\ast_{j}=\emptyset$}
\STATE Add to the mode set, \ie, $\textbf{D}^\ast\leftarrow\textbf{D}^\ast\cup\{\mathcal{D}_{i}\}$
\ELSE
\STATE Remove the overlapping part in the parsed modes, \ie, $\forall j, \mathcal{D}^\ast_{j}\in\textbf{D}^\ast: \hat{\mathcal{D}}_{i}\leftarrow\mathcal{D}_{i}/\mathcal{D}^\ast_{j}$
\STATE Add to the mode set, \ie, $\textbf{D}^\ast\leftarrow\textbf{D}^\ast\cup\{\hat{\mathcal{D}}_{i}\}$
\ENDIF
\ENDFOR
\STATE Obtain the reliable target part set, \ie, $\mathcal{S}=\{p|\forall \nu\in\mathcal{D}^\ast_{i}, \mathcal{D}^\ast_{i}\in\textbf{D}^\ast:p\in\nu\}$
\end{algorithmic}}
\end{algorithm}
\subsection{Reliable Target Parts Based Voting} \label{sec_target_state}
After obtaining $\mathcal{S}$, we determine the target state in the current frame $t$, including center $\ell^t_\ast$ and scale $s^t_\ast$ of the target by reliable target parts based voting. Similar to the method of~\cite{cai2014robust}, we construct a confidence map $C$ to represent location probability of the target in the searching area as
\begin{equation}
\label{equ_confidence_map}
C(i,j)=\left \{
\begin{aligned}
\lambda_1 &\quad (i,j)\in\mathcal{R}_{\mathcal{D}^\ast}\\
\lambda_2 &\quad (i,j)\in\mathcal{R}_{\mathcal{V}^\ast},(i,j)\notin\mathcal{R}_{\mathcal{D}^\ast}\\
\lambda_3 &\quad \text{otherwise}\\
\end{aligned}
\right.
\end{equation}
where $(i,j)$ is the position in the searching area. $\mathcal{R}_{\mathcal{D}^\ast}$ means the region of target parts belonging to the extracted modes, and $\mathcal{R}_{\mathcal{V}^\ast}$ means the region of candidate parts. $\{\lambda_1,\lambda_2,\lambda_3\}$ are constants for the influence of each type of regions.

To find the bounding box to cover more foreground regions with respect to center $\ell$ and scale $s$, we form the following optimization problem
\begin{equation}
\label{equ_calc_location}
\{\ell_\ast^t,s_\ast^t\} = \argmax_{\ell,s}{\sum_{(i,j)\in \mathcal{R}(\ell,s)}{C(i,j)}},
\end{equation}
where $\mathcal{R}(\ell,s)$ means the region with center $\ell$ and scale $s$.

The target center in the current frame $t$ is largely determined by the one in the previous frame $t-1$ with the geometric constraint. To reduce computational complexity, we first estimate a rough target center by calculating the weighted mean of target part center $l^t_p$ with weight $w^t_p$, \ie,
\begin{equation}
\label{equ_calc_center}
\ell^{t} = \sum_{p\in\mathcal{S}}{(\ell^{t-1}_\ast+l_{p}^{t}-l_{p}^{t-1})\cdot \frac{w_{p}^t}{\sum_{p\in\mathcal{S}}{w_{p}^t}}},
\end{equation}
where $\ell^{t-1}_\ast$ is the optimal center in the previous frame $t-1$. $w_{p}^t$ denotes the confidence of the mode including reliable target part $p$ in the current frame $t$, \ie, $w^t_p=\Omega(\mathcal{D}),p\in\nu,\nu\in\mathcal{N}(\mathcal{D})$. After that, we modify the target center with the displacement perturbation term $\delta^t_\ell$ and adjust the target scale with the scale perturbation term $\delta^t_s$ for a visually better location. The maximal values of two perturbation terms $\{\delta^t_\ell,\delta^t_s\}$ are set as the mean diameter of candidate parts in the current frame $t$. The final target state $\{\ell^t_\ast,s^t_\ast\}$ is obtained by optimizing~\eqref{equ_calc_location} using a sampling strategy, namely selecting the one with the maximal score out of numerous randomly sampled states $\{\ell+\delta^t_\ell,s+\delta^t_s\}$. Assembling all parts belonging to the target, we find the optimal target state, as shown in Fig.~\ref{fig_example}(b).
\subsection{Online Updating of Hypergraph}\label{sec_update}
To handle possible significant changes of target appearance, geometric hypergraph $\mathcal{G}$ is updated in two aspects, \ie, target part set $P$ and candidate part set $Q$. As illustrated in Fig.~\ref{fig_example}(a), based on the parsed reliable target part set $\mathcal{S}$, an old part in $P$ (\eg, part $12$) is deleted if it does not involve in any structural correspondence for a fixed number of frames ($5$ frames in the experiment), while a new part (\eg, part $14$ and part $15$) not involved in existed modes is added in $P$ such that its geometric distance to any other parts is larger than a threshold\footnote{The threshold is set as double mean diameter of candidate parts in the current frame.} to preserve the spatial sparsity of $P$. On the other hand, the appearance model in the MRF based segmentation method is updated to generate $Q$ every frame, as similar as in~\cite{cai2014robust}.
\section{Experiments}\label{sec_experiment}
\subsection{Datasets and Protocols}
\subsubsection{VOT2014 dataset}
The VOT2014 dataset~\cite{DBLP:conf/eccv/KristanPLMCNVFL14} is popularly used in the tracking community, which is collected with representative $25$ sequences selected from $394$ sequences. Each sequence is annotated by several attributes such as occlusion, and illumination changes.

We evaluate the tracking methods following two protocols of the VOT2014 challenges, \ie, \textit{Baseline} and \textit{Region\_noise}. \textit{Baseline} corresponds to the experimental setting where the tracker is run on each sequence $15$ times by initializing it on the groundtruth bounding box, obtaining average statistic scores of the measures. \textit{Region\_noise} corresponds to the experiment setting where the tracker is initialized with $15$ noisy bounding boxes, which are randomly perturbed in order of $10\%$ of the groundtruth bounding box size, in each sequence. As defined in~\cite{DBLP:conf/wacv/CehovinKL14}, two performance metrics, \textit{Accuracy} (average bounding box overlap between the bounding box predicted by the tracker and the groundtruth one) and \textit{Robustness} (number of re-initializations once the overlap ratio measure drops to zero) are reported in the experiment.
\subsubsection{Deform-SOT dataset}
To further evaluate the performance of trackers on deformation and occlusion, we collect the Deform-SOT dataset, which includes $50$ challenging sequences and different targets with deformation and occlusion in varying levels in unconstrained environment. The dataset is diverse in object categories, camera viewpoints, sequence lengths and challenging levels. We categorize the difficulty levels of the sequences into six classes, including large deformation, severe occlusion, abnormal movement, illumination variation, scale change and background clutter, for comparison.

We use two popular measures for evaluation, \ie, \textit{precision plot} and \textit{success plot}. The precision plot shows the percentage of successfully tracked frames vs. the center location error in pixels, which ranks the trackers as \textit{precision score} at $20$ pixels; the success plot draws the percentage of successfully tracked frames vs. the bounding box overlap threshold, where Area Under the Curve is used as \textit{success score} for ranking. We run the One-Pass Evaluation (OPE), Spatial Robustness Evaluation (SRE) and Temporal Robustness Evaluation (TRE) for two measures (see definitions in~\cite{wu2013online}) on the dataset.
\subsection{Implementation Details}
The proposed tracker is implemented with MATLAB and C and runs at $0.5$ frame-per-second on a machine with a $2.9$ GHz Intel i7 processor and $16$ GB memory. First of all, we study the influence of several important parameters as follows, where the experiment is performed on $15$ sequences selected from the Deform-SOT dataset with different kinds of challenges.
\subsubsection{Order of Hypergraph} 
\begin{figure}[h]
\centering
\includegraphics[width=.95\linewidth]{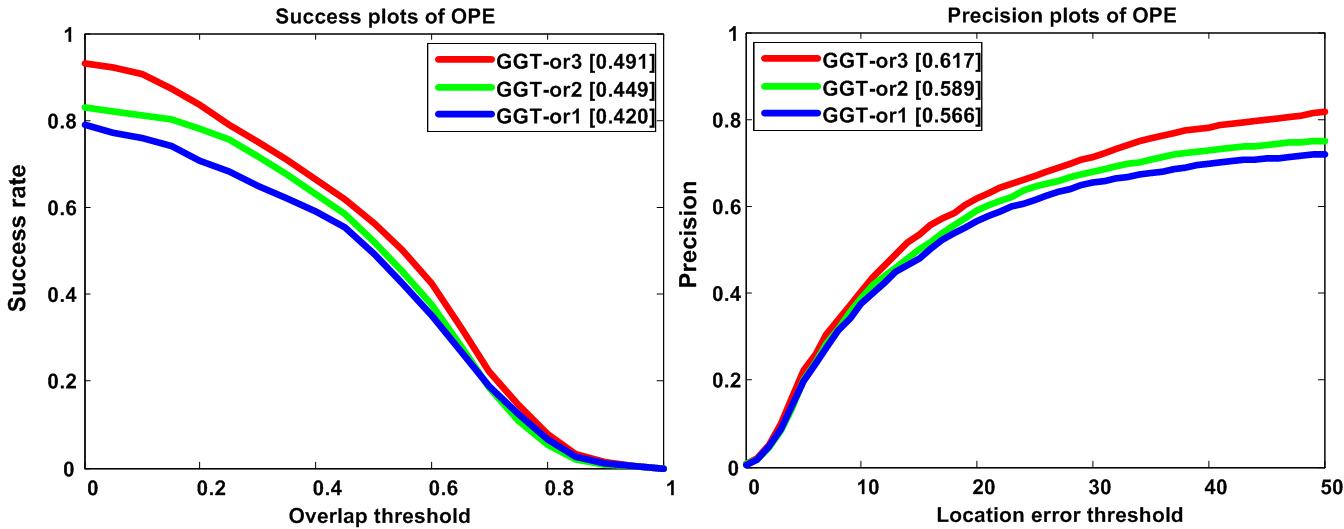}
\caption{Performance vs. order of hypergraph.}
\label{fig_parameter_order}
\end{figure}

The order of hypergraph decides how we consider the geometric relations among correspondence hypotheses. We compare the tracking methods with different orders of hypergraph, denoted as \textrm{GGT-or$k$} ($k=1,2,3$), where the corresponding geometric confidence is calculated in Section~\ref{sec_confidence_calculation}. As shown in Fig.~\ref{fig_parameter_order}, \textrm{GGT-or$3$} considering high-order geometric relations performs the best. In contrast, \textrm{GGT-or$2$} and \textrm{GGT-or$1$} consider just pairwise relations or no relations between parts, leading to big accuracy loss. It indicates the importance and effectiveness of our high-order representation that integrates geometric structural information fully for the target.
\subsubsection{Number of Pixels in Each Superpixel} 
\begin{figure}[h]
\centering
\includegraphics[width=.95\linewidth]{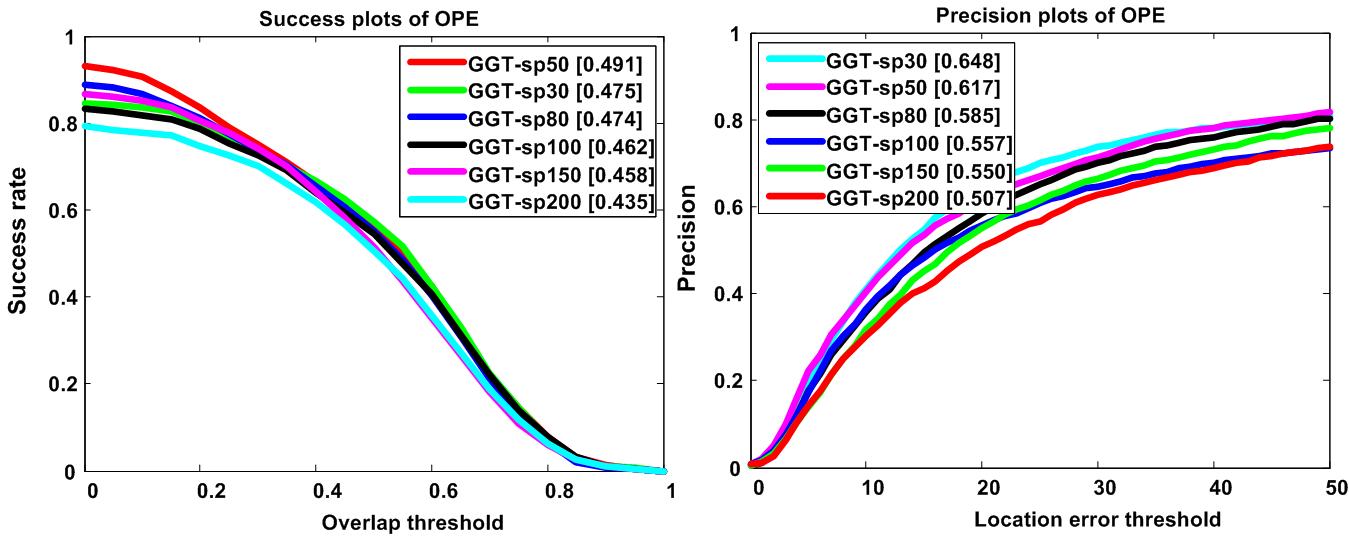}
\caption{Performance vs. number of pixels in superpixel.}
\label{fig_candidate_part}
\end{figure}

The number of superpixel controls the size of parts and the number of vertices in the hypergraph. As Fig.~\ref{fig_candidate_part} shows, we consider different numbers of pixels $\varrho$ in each superpixel, \ie, \textrm{GGT-sp$\varrho$} ($\varrho=30,50,80,100,150,200$). If the number of pixels in each superpixel is too large (\eg, \textrm{GGT-sp200}), it is hard to exploit discriminative geometric structure cues of local parts to handle deformation. On the other hand, if it is too small (\eg, \textrm{GGT-sp30}), the large number of hypotheses increases the computational complexity considerably without apparent performance improvement (\eg, \textrm{GGT-sp30} ranks the second in success score and ranks the first in precision score).
\subsubsection{Weight of Geometric Confidence} 
\begin{figure}[h]
\centering
\includegraphics[width=.95\linewidth]{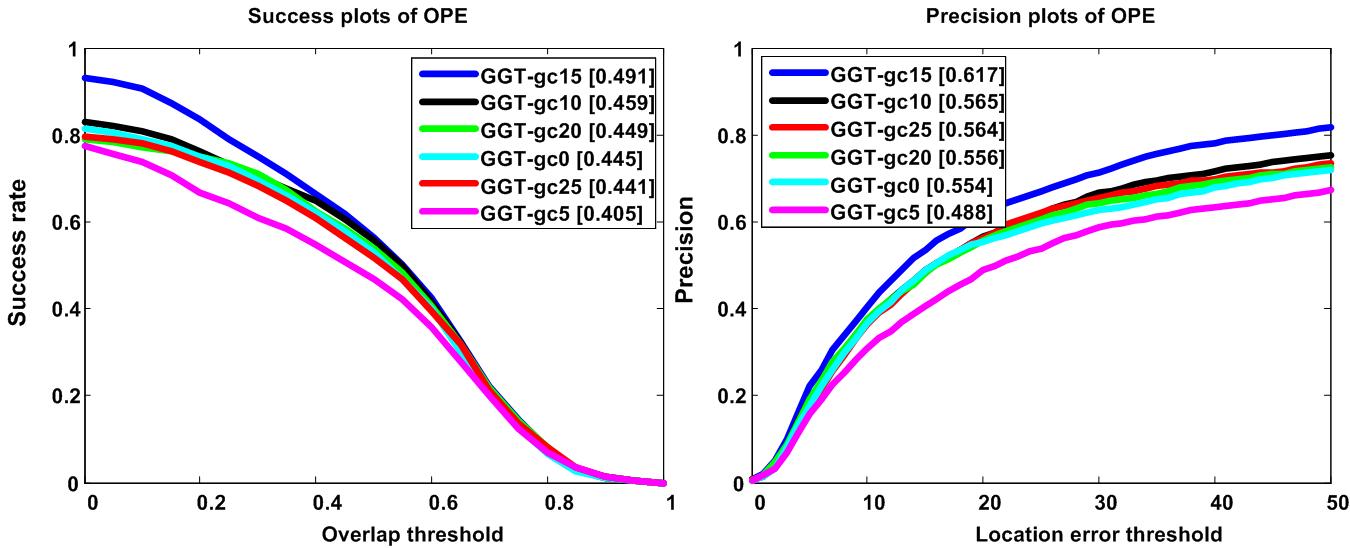}
\caption{Performance vs. weight of geometric confidence.}
\label{fig_graph_confidence}
\end{figure}

The weight of geometric confidence indicates the importance of geometric confidence. Here we set $\omega_1 = 10$ and enumerate the weight $\omega_2$ in~\eqref{equ_association_confidence}, \ie, \textrm{GGT-gc$\omega_2$} ($\omega_2=0,5,10,15,20,25$). Based on the performance in Fig.~\ref{fig_graph_confidence}, an appropriate factor helps the tracker achieve higher performance by neither underestimating nor overestimating the geometric information.
\subsubsection{Maximal Number of Sampled Hyperedges} 
\begin{figure}[h]
\centering
\includegraphics[width=.95\linewidth]{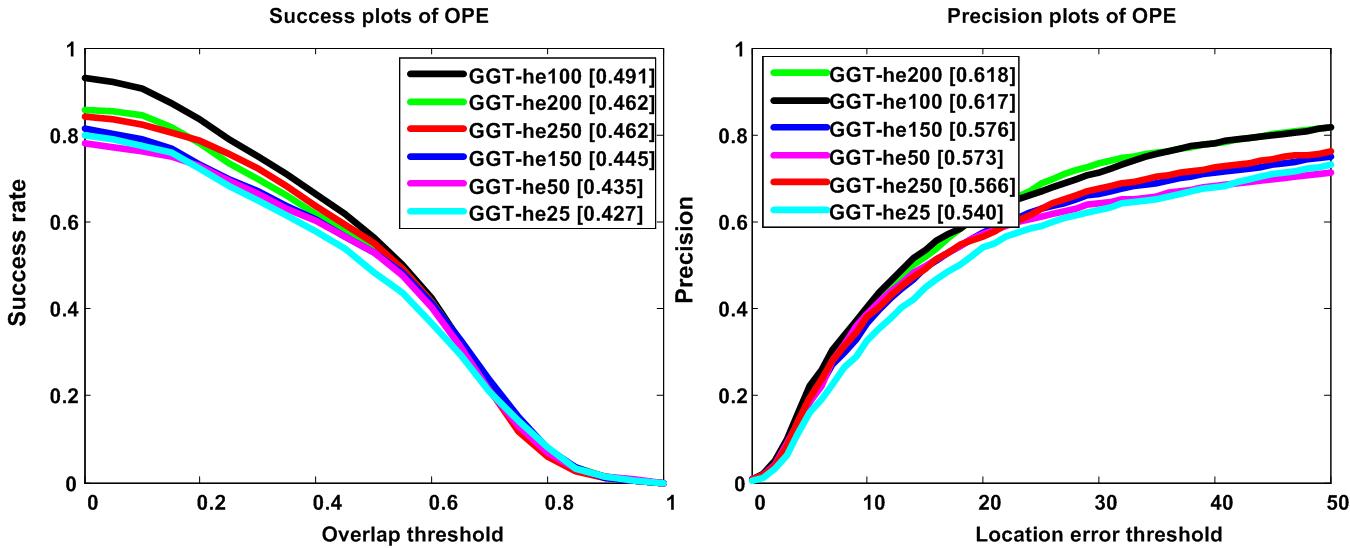}
\caption{Performance vs. number of hyperedges.}
\label{fig_hyper_edges}
\end{figure}

The number of hyperedges measures the importance of geometric information. We report the performance with different numbers of hyperedges in Fig.~\ref{fig_hyper_edges}, denoted as \textrm{GGT-he$N_\nu$} ($N_\nu=25,50,100,150,200,250$). If the number of hyperedges is too small (\eg, \textrm{GGT-he25}), it is insufficient to exploit high-order geometric information, rendering less discriminative structure cues to handle deformation; if the number is too large (\eg, \textrm{GGT-he250}), it is harmful to introduce many noisy relations by the large number of hyperedges because of the sparsity of reliable correspondences.

Based on the above parameter analysis, we set and fix all parameters in our algorithm empirically. The order of the geometric hypergraph is set as $k=3$. For the searching area, we search the target location in current frame by $3$ times the size of previous one. For the SLIC over-segmentation method, the number of pixels in each superpixel is set as $\kappa=50$, and the range of number of superpixels $[100,450]$. We use $8$ bins for each channel of HSV feature to represent the appearance of target parts. The weights in~\eqref{equ_association_confidence} are set as $\{\omega_1,\omega_2\}=\{10,15\}$. The scaling parameters $\sigma_{\nu}^2 = 1.0$ in~\eqref{equ_self_circle}, and $\sigma_{\psi}^2 = 1.0$ in~\eqref{equ_2_affinity}\eqref{equ_3_affinity}. In the sampling method, the appearance threshold is set as $\epsilon_a = 0.3$, and the maximal number of sampled hyperedges is set as $N_\nu=100$. In~\eqref{equ_confidence_map}, the term $\{\lambda_1,\lambda_2,\lambda_3\}=\{3.25,1,-1\}$.
\subsection{Evaluations on the VOT2014 Dataset}\label{sec_VOT}
We compare our approach to several algorithms including the winner of the VOT2014 challenge, DSST~\cite{DBLP:conf/bmvc/DanelljanHKF14}, and two of the top-performing trackers of the online tracking benchmark~\cite{wu2013online}, namely Struck~\cite{hare2011struck} and KCF~\cite{DBLP:journals/pami/HenriquesC0B15}. Furthermore, we include key-point based CMT~\cite{DBLP:conf/wacv/NebehayP14} and IIVTv2~\cite{DBLP:conf/iccv/YiJHCC13}, the part based DGT~\cite{cai2014robust}, LGTv1~\cite{cehovin2013robust}, OGT~\cite{DBLP:conf/eccv/NamHH14}, and PTp~\cite{DBLP:conf/iccv/DuffnerG13}, as well as the baseline trackers including FRT~\cite{adam2006robust}, CT~\cite{zhang2012real}, and MIL~\cite{babenko2011robust}. To ensure a fair comparison, all the results are copied from the original submissions to the VOT2014 challenge by the corresponding authors or the VOT committee.
\begin{figure*}[t]
\centering
\includegraphics[width=.9\linewidth]{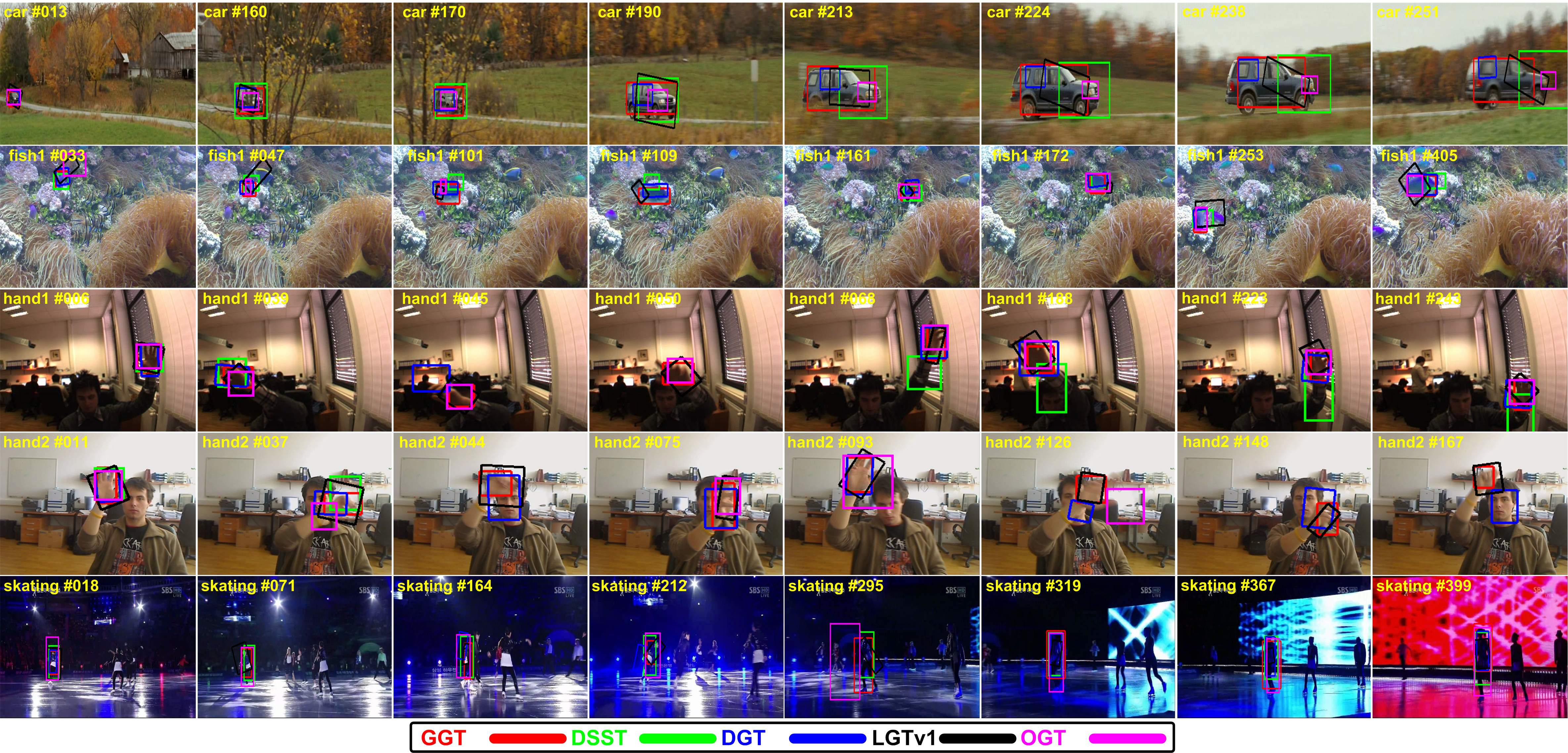}
\caption{Tracking results of $5$ trackers (\ie, GGT, DSST~\cite{DBLP:conf/bmvc/DanelljanHKF14}, DGT~\cite{cai2014robust}, LGTv1~\cite{cehovin2013robust} and OGT~\cite{DBLP:conf/eccv/NamHH14}), are denoted in different colors on the VOT2014 dataset (from top to down are \textit{car}, \textit{fish1}, \textit{hand1}, \textit{hand2}, and \textit{skating}, respectively). Note that one tracker is not shown in some frames, which means it fails in tracking and will re-initialize later (\eg, DSST~\cite{DBLP:conf/bmvc/DanelljanHKF14} fails in \textit{hand1} $\#050$).  Results are best viewed by zooming the digital edition of the figure.}
\label{fig_tracking_result_VOT}
\end{figure*}

{\noindent \textbf{Qualitative Evaluation.}}
Examples of visual tracking results of top $5$ trackers are also shown in Fig.~\ref{fig_tracking_result_VOT}. We can observe that our hypergraph based tracker performs against other graph-based trackers, such as DGT~\cite{cai2014robust}, LGT~\cite{cehovin2013robust}, and OGT~\cite{DBLP:conf/eccv/NamHH14}. For example, DGT~\cite{cai2014robust} and OGT~\cite{DBLP:conf/eccv/NamHH14} do not adjust the scale change of the target \textit{car}. When the figure skater in \textit{skating} moves under the challenges of background clutter and illumination variation, some trackers do not locate well (\eg, OGT~\cite{DBLP:conf/eccv/NamHH14} in $\#295$, and DGT~\cite{cai2014robust} in $\#212$). Besides, DSST~\cite{DBLP:conf/bmvc/DanelljanHKF14} fails in tracking the hand in \textit{hand1} $\#050$ and \textit{hand2} $\#167$. This is because the high-order geometric relations in our method capture the invariant property of local parts such as angles rather than vulnerable pairwise affinity, rendering more tolerance on drastic rotation or appearance variations.

{\noindent \textbf{Quantitative Evaluation.}}
Table~\ref{tab_vot_results} shows the average performance of the compared trackers. As these results show, our algorithm achieves the overall best robustness score, and comparable performance in accuracy among all the methods compared. Moreover, the considerable improvement in \textit{Region\_noise} level indicates that the spatial high-order representation in our method can resist noises effectively, and recover from initialization errors to gain improvements both in terms of accuracy and robustness.
\begin{table*}[t]
\caption{Tracking Results on the VOT2014 dataset. Accuracy scores and ranks (Acc. Sc. and Acc. Rk. for short) are reported as well as the Robustness ones. The first, second and third best values are highlighted by red, blue and green color, respectively.}\label{tab_vot_results}
\centering\begin{tabular}{|c|c|c|c|c|c|c|}
\hline
& \multicolumn{2}{c|}{ \textbf{\textit{Baseline}} }  & \multicolumn{2}{c|}{ \textbf{\textit{Region\_noise}} } & \multicolumn{2}{c|}{ \textbf{Overall} }\\
\hline
&\textbf{Acc. Sc./Acc. Rk.}&\textbf{Rob. Sc./Rob. Rk.}&\textbf{Acc. Sc./Acc. Rk.}&\textbf{Rob. Sc./Rob. Rk.}&\textbf{Acc. Sc./Acc. Rk.}&\textbf{Rob. Sc./Rob. Rk.}\\\hline
GGT&0.58/6.16&{\color{red}0.55}/{\color{red}4.98} &0.57/{\color{green}4.81}&{\color{blue}0.65}/{\color{red}4.93} &0.57/5.48&{\color{red}0.59}/{\color{red}4.95}\\\hline
DSST~\cite{DBLP:conf/bmvc/DanelljanHKF14}&{\color{blue} 0.62}/{\color{blue}4.48}&1.16/6.32 &{\color{red}0.58}/{\color{red}4.01}&1.28/6.22 &{\color{blue} 0.60}/{\color{red}4.25}&1.22/6.27\\\hline
DGT~\cite{cai2014robust}&{\color{green} 0.58}/{\color{green}5.81}&{\color{green}1.00}/{\color{blue}5.02} &{\color{green}0.58}/4.97&{\color{green}1.17}/{\color{blue}5.31} &{\color{green}0.58}/{\color{green}5.39}&{\color{green}1.09}/{\color{blue}5.16}\\\hline
KCF~\cite{DBLP:journals/pami/HenriquesC0B15}&{\color{red} 0.63}/{\color{red}4.22}&1.32/6.53 &{\color{blue}0.58}/{\color{blue}4.50}&1.52/6.62 &{\color{red}0.61}/{\color{blue}4.36}&1.42/6.57\\\hline
LGTv1~\cite{cehovin2013robust}&0.47/9.29&{\color{blue}0.66}/{\color{green}5.96} &0.46/8.73&{\color{red}0.64}/{\color{green}5.42} &0.47/9.01&{\color{blue} 0.65}/{\color{green}5.69}\\\hline
Struck~\cite{hare2011struck}&0.52/8.04&2.16/8.64 &0.49/7.90&2.22/8.16 &0.51/7.97&2.19/8.40\\\hline
OGT~\cite{DBLP:conf/eccv/NamHH14}&0.55/7.09&3.34/9.78 &0.51/7.19&3.37/10.30 &0.53/7.14&3.36/10.04\\\hline
PTp~\cite{DBLP:conf/iccv/DuffnerG13}&0.47/10.98&1.40/7.20 &0.45/9.77&1.46/7.33 &0.46/10.38&1.43/7.26\\\hline
CMT~\cite{DBLP:conf/wacv/NebehayP14}&0.48/9.18&2.64/9.16 &0.44/9.97&2.64/9.14 &0.46/9.58&2.64/9.15\\\hline
FoT~\cite{wendel2011robustifying}&0.51/8.44&2.28/9.69 &0.48/9.13&2.71/10.59 &0.50/8.79&2.50/10.14\\\hline
IIVTv2~\cite{DBLP:conf/iccv/YiJHCC13}&0.47/9.30&3.19/9.70 &0.45/9.96&3.13/9.14 &0.46/9.63&3.16/9.42\\\hline
FSDT~\cite{DBLP:conf/eccv/KristanPLMCNVFL14}&0.47/9.87&3.08/11.26 &0.46/9.36&2.77/10.38 &0.47/9.62&2.93/10.82\\\hline
IVT~\cite{lim2004incremental}&0.47/9.87&2.76/10.44 &0.44/10.69&2.86/10.20 &0.46/10.28&2.81/10.32\\\hline
CT~\cite{zhang2012real}&0.43/11.76&3.12/10.23 &0.43/11.04&3.34/10.45 &0.43/11.40&3.23/10.34\\\hline
FRT~\cite{adam2006robust}&0.48/9.17&3.32/12.20 &0.44/10.20&3.46/12.29 &0.46/9.69&3.39/12.24\\\hline
MIL~\cite{babenko2011robust}&0.40/12.03&2.27/8.80 &0.35/13.67&2.60/9.67 &0.38/12.85&2.44/9.23\\\hline
\end{tabular}
\end{table*}
\subsection{Evaluations on the Deform-SOT dataset}\label{sec_SOT}
We evaluate the proposed algorithm against exsiting methods including holistic model based trackers (\ie, IVT~\cite{lim2004incremental}, L1T~\cite{mei2009robust}, TLD~\cite{kalal2010pn}, MIL~\cite{babenko2011robust}, Struck~\cite{hare2011struck}, MTT~\cite{zhang2012robust}, CT~\cite{zhang2012real}, CN~\cite{danelljan2014adaptive}, STT~\cite{wen2012online} and STC~\cite{zhang2014fast}) and part based trackers (\ie, Frag~\cite{adam2006robust}, SPT~\cite{wang2011superpixel}, SCM~\cite{DBLP:conf/cvpr/ZhongLY12}, LOT~\cite{oron2012locally}, ASLA~\cite{DBLP:conf/cvpr/JiaLY12}, LSL~\cite{yao2013part}, LGT~\cite{cehovin2013robust}, DGT~\cite{cai2014robust}, and TCP~\cite{li2015online}). For fair comparison, we use the {\em same} initial bounding box of each sequence for all trackers. The experimental results of other trackers are reproduced from the available source codes with recommended parameters.

\begin{figure*}[ht]
\centering
\includegraphics[width=.95\linewidth]{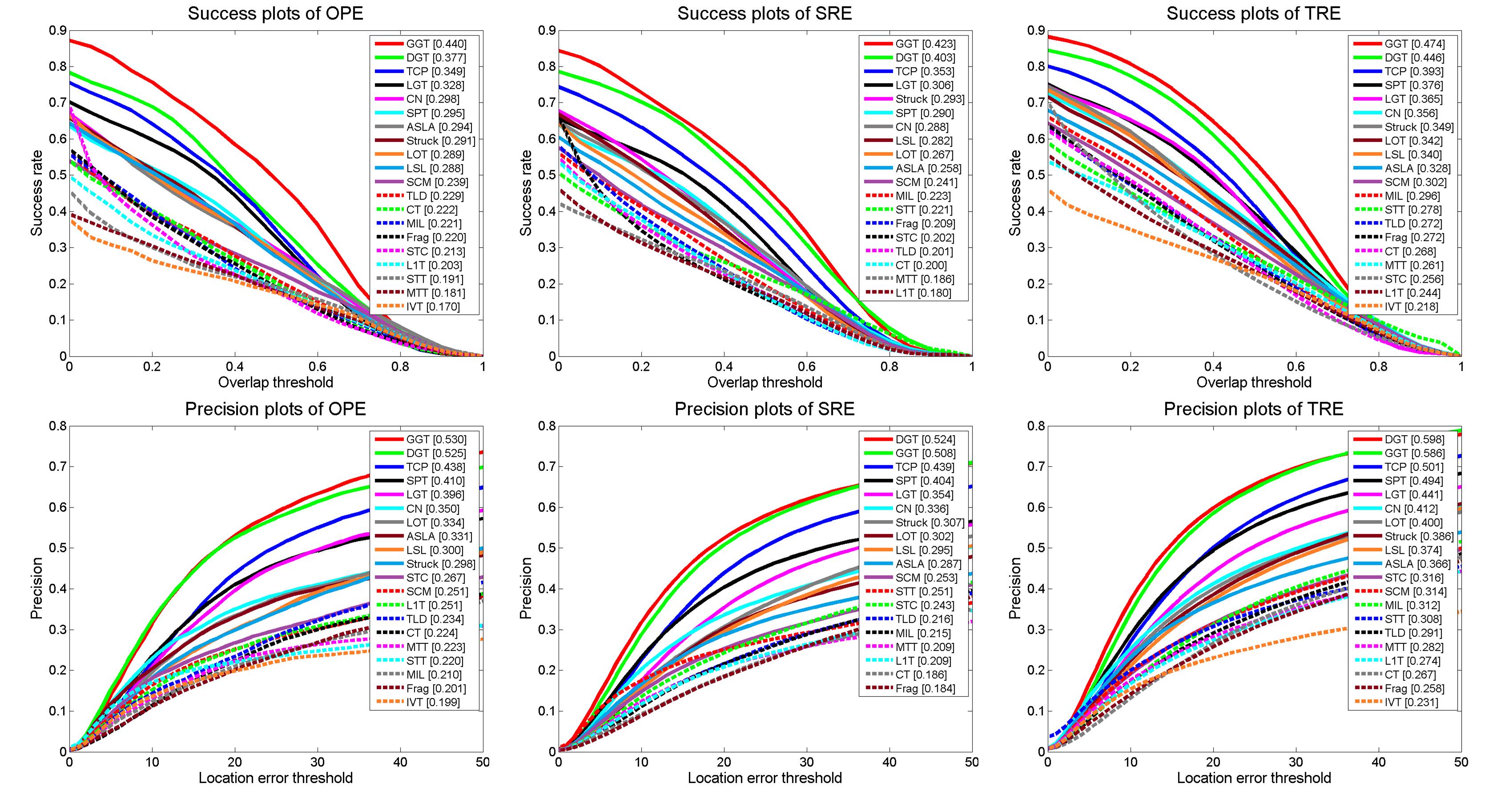}
\caption{Precision plot and success plot over the Deform-SOT dataset using OPE, SRE and TRE. Best viewed in color.}
\label{fig_overlapthreshol_plot}
\end{figure*}

As shown in Fig.~\ref{fig_overlapthreshol_plot}, the evaluation results on OPE, SRE and TRE indicate that our GGT tracker performs against other compared methods. In addition, Fig.~\ref{fig_tracking_result_SOT} shows the tracking results of top $5$ trackers on several sequences.

\begin{figure*}[ht]
\centering
\includegraphics[width=.95\linewidth]{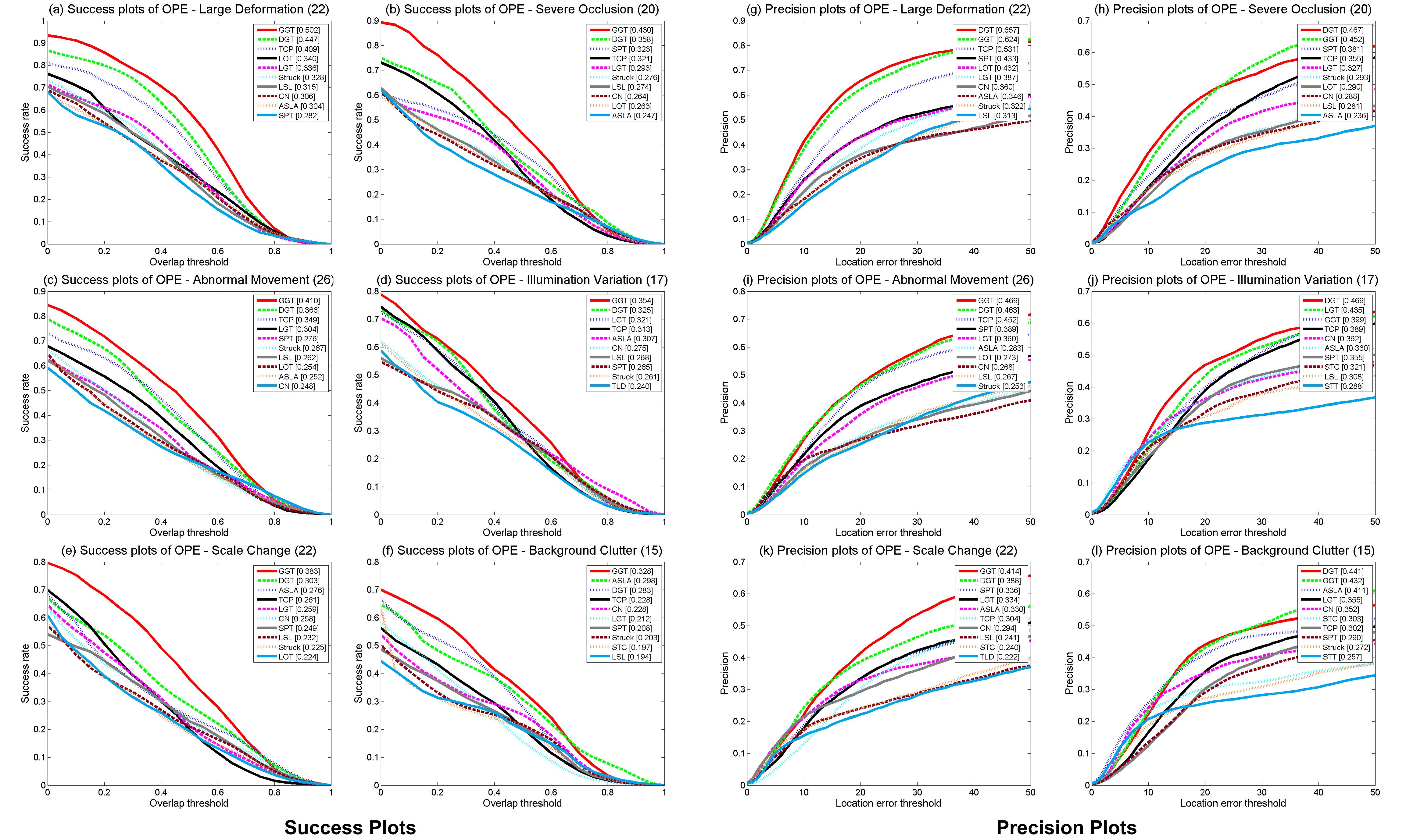}
\caption{The plots of OPE with different attributes. Best viewed in color. For better clarity, the top $10$ trackers are shown.}
\label{fig_OPE}
\end{figure*}

{\noindent \textbf{Attribute-based Evaluation.}}We also compare the performance of all tracking algorithms for videos with varying degrees of $6$ challenging factors shown in Fig.~\ref{fig_OPE}.
\subsubsection{Large Deformation}
Existing part based trackers~\cite{wang2011superpixel, cehovin2013robust, yao2013part, cai2014robust} mainly consider vulnerable pairwise geometric relations, which are prone to fail in the sequences with significant target deformation (\eg, \textit{boarding} in Fig.~\ref{fig_tracking_result_SOT}). According to Fig.~\ref{fig_OPE}(a)(g), our tracker performs against other methods because high-order triangle geometric relations instead of varying pairwise displacements preserve invariant angles to remove noises from a large set of correspondence hypotheses.
\subsubsection{Severe Occlusion}
Some trackers~\cite{mei2009robust, kalal2010pn, wang2011superpixel, DBLP:conf/cvpr/ZhongLY12, DBLP:conf/cvpr/JiaLY12, yao2013part} drift away from the target or do not scale well when the target is heavily occluded (\eg, \textit{boarding}, \textit{carscale}, \textit{run} and \textit{waterski} in Fig.~\ref{fig_tracking_result_SOT}). However, our method is able to track the target relatively accurate because the structural correspondence modes exploit invariant local geometric structure of target parts. This information helps to avoid much influence of occlusion as long as adequate structural correspondence modes are detected to parse target parts to vote for the target state.
\subsubsection{Abnormal Movement}
Abnormal movements consist of all kinds of non-rigid change such as abrupt motion, pose variation, and rotation. For example, SCM~\cite{DBLP:conf/cvpr/ZhongLY12} and TCP~\cite{li2015online} drift away when the gymnast jumps to grab bars in~\textit{uneven-bars} $\#303$. By comparison, our method performs well in estimating both scales and positions on these challenging sequences, which can be attributed to two reasons. Firstly, the hypergraph is constructed with coarse target parts to remove unnecessary background parts (see a example in Fig.~\ref{fig_example}(a)). Moreover, based on the modes, the reliable target parts are determined under noises to vote the optimal target state.
\subsubsection{Illumination Variation}
Some trackers~\cite{DBLP:conf/cvpr/JiaLY12, danelljan2014adaptive} are insensitive to appearance changes caused by illumination variation, however, compared to our method, they perform poorly on the sequences undergoing other challenges such as large deformation and abnormal movement simultaneously (see \textit{bike} in Fig.~\ref{fig_tracking_result_SOT}). This can be attributed to the use of geometric hypergraph learning to adapt the local parts' appearance variation in consecutive frames.
\subsubsection{Scale Change}
In terms of sequences with significant target scale change (\eg, \textit{boarding} and \textit{carscale} in Fig.~\ref{fig_tracking_result_SOT}), our tracker performs against other methods~\cite{DBLP:conf/cvpr/JiaLY12, cai2014robust, li2015online} as in Fig.~\ref{fig_OPE}(e)(k). This is because we employ the angles of the triangle to measure the similarity of several correspondence hypotheses, which is invariant to scale change (see more in Section~\ref{sec_sampling}). Different from our algorithm, DGT~\cite{cai2014robust} just considers neighboring pairwise relations between local parts, making it less flexible to handle changes of the target scale.
\subsubsection{Background Clutter}
\begin{figure*}[!htp]
\centering
\includegraphics[width=.95\linewidth]{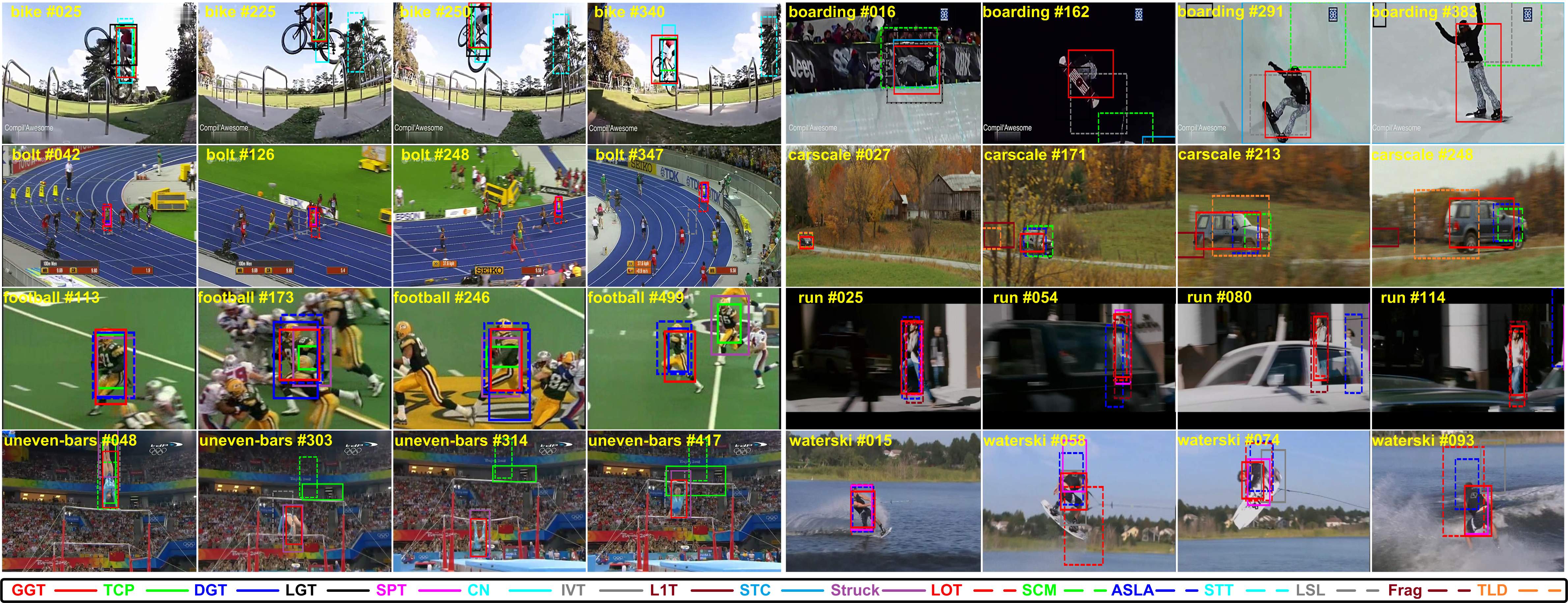}
\caption{Tracking results of top $5$ trackers, denoted in different colors and lines, on the Deform-SOT dataset (from left to right and top to down are \textit{bike}, \textit{boarding}, \textit{bolt}, \textit{carscale}, \textit{football}, \textit{run}, \textit{uneven-bars}, and \textit{waterski}, respectively). Results are best viewed by zooming the digital edition of the figure.}
\label{fig_tracking_result_SOT}
\end{figure*}

The background surrounding the target has similar appearance, leading to drift from the intended target to other objects when they appear in close proximity (\eg, \textit{football} $\#499$ in Fig.~\ref{fig_tracking_result_SOT}). To handle this problem, some methods~\cite{wen2012online, zhang2014fast} exploit the context information around the target, while the other ones~\cite{cehovin2013robust, cai2014robust} employ a graph based representation to capture geometric structure of the target. Owing to the proposed confidence-aware sampling method without distance constraint, sampled representative hyperedges not only consider the relations between target parts and background parts (\textit{context}), but also model the inlier geometric relations among local target parts (\textit{structure}) simultaneously. As a whole, our method ranks the first in success score in Fig.~\ref{fig_OPE}(f) and the second in precision score in Fig.~\ref{fig_OPE}(i).
\section{Conclusion and Future Work}\label{sec_conclusion}
In this paper, we describe the Geometric hyperGraph Tracker (GGT) based on geometric hypergraph learning for visual tracking, where $k$-order geometric relations among {\em correspondence hypotheses} are integrated in the dynamically constructed {\em geometric hypergraph}. Our method is universal in that the traditional graph-based tracking methods can be viewed as special cases of the proposed algorithm with lower-order of hypergraph. On the other hand, the confidence-aware sampling method is developed to reduce computational complexity and the scale of hypergraph for better efficiency. Experiments are carried out in the VOT2014 dataset and the Deform-SOT dataset, which include large deformation and severe occlusion challenges, to demonstrate the favorable performance of the proposed method compared to other existing methods.

There are some issues in our method which can be further improved in the future work. To characterize more kinds of graph-based trackers, we can exploit high-order temporal and spatial relations among a large number of correspondence hypotheses in multiple consecutive frames simultaneously. More spatio-temporal relations considered in the model means higher computational complexity. Therefore, one future direction is to introduce an more effective
mechanism for selecting vertices and hyperedges to reduce redundant hypotheses. Another possible direction is to learn a holistic target representation which is updated jointly with the simple superpixel representation for more robustness and discriminability.

\ifCLASSOPTIONcaptionsoff
  \newpage
\fi

\bibliographystyle{ieee}
\bibliography{reference}

\begin{thebibliography}{10}\itemsep=-1pt

\bibitem{achanta2012slic}
R.~Achanta, A.~Shaji, K.~Smith, A.~Lucchi, P.~Fua, and S.~Susstrunk.
\newblock Slic superpixels compared to state-of-the-art superpixel methods.
\newblock {\em IEEE Transactions on Pattern Analysis and Machine Intelligence},
  34(11):2274--2282, 2012.

\bibitem{adam2006robust}
A.~Adam, E.~Rivlin, and I.~Shimshoni.
\newblock Robust fragments-based tracking using the integral histogram.
\newblock In {\em Proceedings of IEEE Conference on Computer Vision and Pattern
  Recognition}, volume~1, pages 798--805, 2006.

\bibitem{babenko2011robust}
B.~Babenko, M.-H. Yang, and S.~Belongie.
\newblock Robust object tracking with online multiple instance learning.
\newblock {\em IEEE Transactions on Pattern Analysis and Machine Intelligence},
  33(8):1619--1632, 2011.

\bibitem{DBLP:journals/cviu/BouachirB15}
W.~Bouachir and G.~Bilodeau.
\newblock Collaborative part-based tracking using salient local predictors.
\newblock {\em Computer Vision and Image Understanding}, 137:88--101, 2015.

\bibitem{DBLP:journals/pami/BoykovK04}
Y.~Boykov and V.~Kolmogorov.
\newblock An experimental comparison of min-cut/max-flow algorithms for energy
  minimization in vision.
\newblock {\em IEEE Transactions on Pattern Analysis and Machine Intelligence},
  26(9):1124--1137, 2004.

\bibitem{cai2014robust}
Z.~Cai, L.~Wen, Z.~Lei, N.~Vasconcelos, and S.~Z. Li.
\newblock Robust deformable and occluded object tracking with dynamic graph.
\newblock {\em IEEE Transactions on Image Processing}, 23(12):5497--5509, 2014.

\bibitem{cehovin2013robust}
L.~Cehovin, M.~Kristan, and A.~Leonardis.
\newblock Robust visual tracking using an adaptive coupled-layer visual model.
\newblock {\em IEEE Transactions on Pattern Analysis and Machine Intelligence},
  35(4):941--953, 2013.

\bibitem{DBLP:conf/wacv/CehovinKL14}
L.~Cehovin, M.~Kristan, and A.~Leonardis.
\newblock Is my new tracker really better than yours?
\newblock In {\em IEEE Winter Conference on Applications of Computer Vision},
  pages 540--547, 2014.

\bibitem{DBLP:conf/bmvc/DanelljanHKF14}
M.~Danelljan, G.~H{\"{a}}ger, F.~S. Khan, and M.~Felsberg.
\newblock Accurate scale estimation for robust visual tracking.
\newblock In {\em Proceedings of British Machine Vision Conference}, 2014.

\bibitem{danelljan2014adaptive}
M.~Danelljan, F.~Shahbaz~Khan, M.~Felsberg, and J.~Van~de Weijer.
\newblock Adaptive color attributes for real-time visual tracking.
\newblock In {\em Proceedings of IEEE Conference on Computer Vision and Pattern
  Recognition}, 2014.

\bibitem{DBLP:conf/iccv/DuffnerG13}
S.~Duffner and C.~Garcia.
\newblock Pixeltrack: {A} fast adaptive algorithm for tracking non-rigid
  objects.
\newblock In {\em Proceedings of the IEEE International Conference on Computer
  Vision}, pages 2480--2487, 2013.

\bibitem{DBLP:conf/eccv/KristanPLMCNVFL14}
M.~K. et~al.
\newblock The visual object tracking {VOT2014} challenge results.
\newblock In {\em Workshops in Conjunction with European Conference on Computer
  Vision}, pages 191--217, 2014.

\bibitem{DBLP:conf/iccv/GodecRB11}
M.~Godec, P.~M. Roth, and H.~Bischof.
\newblock Hough-based tracking of non-rigid objects.
\newblock In {\em Proceedings of the IEEE International Conference on Computer
  Vision}, pages 81--88, 2011.

\bibitem{DBLP:journals/cviu/GuoCTLLL14}
Y.~Guo, Y.~Chen, F.~Tang, A.~Li, W.~Luo, and M.~Liu.
\newblock Object tracking using learned feature manifolds.
\newblock {\em Computer Vision and Image Understanding}, 118:128--139, 2014.

\bibitem{hare2011struck}
S.~Hare, A.~Saffari, and P.~H. Torr.
\newblock Struck: Structured output tracking with kernels.
\newblock In {\em Proceedings of the IEEE International Conference on Computer
  Vision}, pages 263--270, 2011.

\bibitem{DBLP:conf/cvpr/HareST12}
S.~Hare, A.~Saffari, and P.~H.~S. Torr.
\newblock Efficient online structured output learning for keypoint-based object
  tracking.
\newblock In {\em Proceedings of IEEE Conference on Computer Vision and Pattern
  Recognition}, pages 1894--1901, 2012.

\bibitem{DBLP:journals/pami/HenriquesC0B15}
J.~F. Henriques, R.~Caseiro, P.~Martins, and J.~Batista.
\newblock High-speed tracking with kernelized correlation filters.
\newblock {\em IEEE Transactions on Pattern Analysis and Machine Intelligence},
  37(3):583--596, 2015.

\bibitem{DBLP:conf/eccv/HongWMPT14}
Z.~Hong, C.~Wang, X.~Mei, D.~Prokhorov, and D.~Tao.
\newblock Tracking using multilevel quantizations.
\newblock In {\em European Conference on Computer Vision}, volume 8694, pages
  155--171, 2014.

\bibitem{DBLP:conf/cvpr/JiaLY12}
X.~Jia, H.~Lu, and M.~Yang.
\newblock Visual tracking via adaptive structural local sparse appearance
  model.
\newblock In {\em Proceedings of IEEE Conference on Computer Vision and Pattern
  Recognition}, pages 1822--1829, 2012.

\bibitem{kalal2010pn}
Z.~Kalal, J.~Matas, and K.~Mikolajczyk.
\newblock {P-N} learning: Bootstrapping binary classifiers by structural
  constraints.
\newblock In {\em Proceedings of IEEE Conference on Computer Vision and Pattern
  Recognition}, pages 49--56, 2010.

\bibitem{DBLP:conf/cvpr/LeeCL11}
J.~Lee, M.~Cho, and K.~M. Lee.
\newblock Hyper-graph matching via reweighted random walks.
\newblock In {\em Proceedings of IEEE Conference on Computer Vision and Pattern
  Recognition}, pages 1633--1640, 2011.

\bibitem{li2015online}
W.~Li, L.~Wen, M.~C. Chuah, Y.~Zhang, Z.~Lei, and S.~Z. Li.
\newblock Online visual tracking using temporally coherent part cluster.
\newblock In {\em IEEE Winter Conference on Applications of Computer Vision},
  pages 9--16, 2015.

\bibitem{lim2004incremental}
J.~Lim, D.~A. Ross, R.-S. Lin, and M.-H. Yang.
\newblock Incremental learning for visual tracking.
\newblock In {\em Advances in Neural Information Processing Systems}, pages
  793--800, 2004.

\bibitem{liu2012dense}
H.~Liu, X.~Yang, L.~J. Latecki, and S.~Yan.
\newblock Dense neighborhoods on affinity graph.
\newblock {\em International Journal of Computer Vision}, 98(1):65--82, 2012.

\bibitem{DBLP:conf/cvpr/LiuWY15}
T.~Liu, G.~Wang, and Q.~Yang.
\newblock Real-time part-based visual tracking via adaptive correlation
  filters.
\newblock In {\em Proceedings of IEEE Conference on Computer Vision and Pattern
  Recognition}, pages 4902--4912, 2015.

\bibitem{mei2009robust}
X.~Mei and H.~Ling.
\newblock Robust visual tracking using $\ell1$ minimization.
\newblock In {\em Proceedings of the IEEE International Conference on Computer
  Vision}, pages 1436--1443, 2009.

\bibitem{DBLP:conf/eccv/NamHH14}
H.~Nam, S.~Hong, and B.~Han.
\newblock Online graph-based tracking.
\newblock In {\em European Conference on Computer Vision}, pages 112--126,
  2014.

\bibitem{nebehay2015clustering}
G.~Nebehay and R.~Pflugfelder.
\newblock Clustering of static-adaptive correspondences for deformable object
  tracking.
\newblock In {\em Proceedings of the IEEE Conference on Computer Vision and
  Pattern Recognition}, pages 2784--2791, 2015.

\bibitem{DBLP:conf/wacv/NebehayP14}
G.~Nebehay and R.~P. Pflugfelder.
\newblock Consensus-based matching and tracking of keypoints for object
  tracking.
\newblock In {\em Winter Conference on Applications of Computer Vision}, pages
  862--869, 2014.

\bibitem{oron2012locally}
S.~Oron, A.~Bar-Hillel, D.~Levi, and S.~Avidan.
\newblock Locally orderless tracking.
\newblock In {\em Proceedings of IEEE Conference on Computer Vision and Pattern
  Recognition}, pages 1940--1947, 2012.

\bibitem{ren2007tracking}
X.~Ren and J.~Malik.
\newblock Tracking as repeated figure/ground segmentation.
\newblock In {\em Proceedings of IEEE Conference on Computer Vision and Pattern
  Recognition}, pages 1--8, 2007.

\bibitem{DBLP:journals/tcyb/WangY14b}
J.~Wang and Y.~Yagi.
\newblock Many-to-many superpixel matching for robust tracking.
\newblock {\em {IEEE} Transactions on Cybernetics}, 44(7):1237--1248, 2014.

\bibitem{DBLP:journals/tsmc/WangCX11}
Q.~Wang, F.~Chen, and W.~Xu.
\newblock Tracking by third-order tensor representation.
\newblock {\em {IEEE} Transactions on Systems, Man, and Cybernetics, Part {B}},
  41(2):385--396, 2011.

\bibitem{wang2011superpixel}
S.~Wang, H.~Lu, F.~Yang, and M.-H. Yang.
\newblock Superpixel tracking.
\newblock In {\em Proceedings of the IEEE International Conference on Computer
  Vision}, pages 1323--1330, 2011.

\bibitem{DBLP:conf/accv/WangN12}
W.~Wang and R.~Nevatia.
\newblock Robust object tracking using constellation model with superpixel.
\newblock In {\em Asian Conference on Computer Vision}, pages 191--204, 2012.

\bibitem{DBLP:conf/accv/WenCDLL14}
L.~Wen, Z.~Cai, D.~Du, Z.~Lei, and S.~Z. Li.
\newblock Learning discriminative hidden structural parts for visual tracking.
\newblock In {\em Workshops in Conjunction with Asian Conference on Computer
  Vision}, pages 262--276, 2014.

\bibitem{wen2012online}
L.~Wen, Z.~Cai, Z.~Lei, D.~Yi, and S.~Z. Li.
\newblock Online spatio-temporal structural context learning for visual
  tracking.
\newblock In {\em European Conference on Computer Vision}, pages 716--729,
  2012.

\bibitem{wen2015jots}
L.~Wen, D.~Du, Z.~Lei, S.~Z. Li, and M.-H. Yang.
\newblock {JOTS}: Joint online tracking and segmentation.
\newblock In {\em Proceedings of IEEE Conference on Computer Vision and Pattern
  Recognition}, pages 2226--2234, 2015.

\bibitem{wendel2011robustifying}
A.~Wendel, S.~Sternig, and M.~Godec.
\newblock Robustifying the flock of trackers.
\newblock In {\em Computer Vision Winter Workshop}, page~91. Citeseer, 2011.

\bibitem{wu2013online}
Y.~Wu, J.~Lim, and M.-H. Yang.
\newblock Online object tracking: A benchmark.
\newblock In {\em Proceedings of IEEE Conference on Computer Vision and Pattern
  Recognition}, pages 2411--2418, 2013.

\bibitem{DBLP:journals/ivc/YangLY13}
F.~Yang, H.~Lu, and M.~Yang.
\newblock Learning structured visual dictionary for object tracking.
\newblock {\em Image Vision Computing}, 31(12):992--999, 2013.

\bibitem{yao2013part}
R.~Yao, Q.~Shi, C.~Shen, Y.~Zhang, and A.~van~den Hengel.
\newblock Part-based visual tracking with online latent structural learning.
\newblock In {\em Proceedings of IEEE Conference on Computer Vision and Pattern
  Recognition}, pages 2363--2370, 2013.

\bibitem{DBLP:conf/iccv/YiJHCC13}
K.~M. Yi, H.~Jeong, B.~Heo, H.~J. Chang, and J.~Y. Choi.
\newblock Initialization-insensitive visual tracking through voting with
  salient local features.
\newblock In {\em Proceedings of the IEEE International Conference on Computer
  Vision}, pages 2912--2919, 2013.

\bibitem{DBLP:journals/tcyb/YuYWH15}
X.~Yu, J.~Yang, T.~Wang, and T.~S. Huang.
\newblock Key point detection by max pooling for tracking.
\newblock {\em {IEEE} Transactions on Cybernetics}, 45(3):444--452, 2015.

\bibitem{zhang2014fast}
K.~Zhang, L.~Zhang, Q.~Liu, D.~Zhang, and M.~Yang.
\newblock Fast visual tracking via dense spatio-temporal context learning.
\newblock In {\em European Conference on Computer Vision}, pages 127--141,
  2014.

\bibitem{zhang2012real}
K.~Zhang, L.~Zhang, and M.-H. Yang.
\newblock Real-time compressive tracking.
\newblock In {\em European Conference on Computer Vision}, pages 864--877.
  Springer, 2012.

\bibitem{zhang2012robust}
T.~Zhang, B.~Ghanem, S.~Liu, and N.~Ahuja.
\newblock Robust visual tracking via multi-task sparse learning.
\newblock In {\em Proceedings of IEEE Conference on Computer Vision and Pattern
  Recognition}, pages 2042--2049, 2012.

\bibitem{DBLP:conf/cvpr/ZhongLY12}
W.~Zhong, H.~Lu, and M.~Yang.
\newblock Robust object tracking via sparsity-based collaborative model.
\newblock In {\em Proceedings of IEEE Conference on Computer Vision and Pattern
  Recognition}, pages 1838--1845, 2012.

\end{thebibliography}
\end{document}